%% file: main.tex
\pdfoutput=1
\documentclass[11pt]{article}

\usepackage[]{acl}

\usepackage{times}
\usepackage{latexsym}

\usepackage[T1]{fontenc}
\usepackage[utf8]{inputenc}

\usepackage{microtype}
\usepackage{comment}
\usepackage{times}
\usepackage{latexsym}
\usepackage{graphicx}
\usepackage{amsmath}
\usepackage{amssymb}
\usepackage{booktabs}
\usepackage{textcomp}
\usepackage{multirow}
\usepackage{array}
\usepackage{bbm}
\usepackage{tabularx, booktabs}
\newcolumntype{Y}{>{\centering\arraybackslash}X}
\newcolumntype{L}{>{\arraybackslash}X}

\usepackage{tablefootnote}
\usepackage{subcaption}
\usepackage{hyperref}
\usepackage{makecell}

\usepackage[T1]{fontenc}
\usepackage[utf8]{inputenc}

\usepackage{enumitem}
\usepackage{microtype}
\usepackage{enumitem}
\usepackage{xparse}
\definecolor{bluepigment}{rgb}{0.2, 0.2, 0.6}
\definecolor{emerald}{rgb}{0.31, 0.78, 0.47}
\definecolor{orange}{rgb}{0.91, 0.41, 0.17}
\newcommand{\ours}{Our Method}

\ifx\nocomment\undefined
    \NewDocumentCommand{\heng}
    { mO{} }{\textcolor{red}{\textsuperscript{\textit{Heng}}\textsf{\textbf{\small[#1]}}}}
    \NewDocumentCommand{\jh}
    { mO{} }{\textcolor{purple}{\textsuperscript{\textit{Jiawei}}\textsf{\textbf{\small[#1]}}}}
    \NewDocumentCommand{\zoey}
    { mO{} }{\textcolor{bluepigment}{\textsuperscript{\textit{Zoey}}\textsf{\textbf{\small[#1]}}}}
    \NewDocumentCommand{\qingyun}
    { mO{} }{\textcolor{cyan}{\textsuperscript{\textit{qingyun}}\textsf{\textbf{\small[#1]}}}}
    \NewDocumentCommand{\yi}
    { mO{} }{\textcolor{green}{\textsuperscript{\textit{yi}}\textsf{\textbf{\small[#1]}}}}
    \NewDocumentCommand{\revanth}
    { mO{} }{\textcolor{orange}{\textsuperscript{\textit{Revanth}}\textsf{\textbf{\small[#1]}}}}
    \NewDocumentCommand{\clare}
    { mO{} }{\textcolor{teal}{\textsuperscript{\textit{Clare}}\textsf{\textit{\small[#1]}}}}

\else
    \NewDocumentCommand{\heng}
    { mO{} }{\textcolor{red}{}}
    \NewDocumentCommand{\jh}
    { mO{} }{\textcolor{purple}{}}
    \NewDocumentCommand{\zoey}
    { mO{} }{\textcolor{bluepigment}{}}
\fi

\title{Schema-Guided Culture-Aware Complex Event Simulation \\ with Multi-Agent Role-Play %\\
}

\author{Sha Li$^1$, Revanth Gangi Reddy$^1$, Khanh Duy Nguyen$^1$, Qingyun Wang$^1$, May Fung$^1$,\\\textbf{Chi Han$^1$, Jiawei Han$^1$, Kartik Natarajan$^2$, Clare R. Voss$^3$, Heng Ji$^1$}\\
$^1$University of Illinois Urbana-Champaign \\
$^2$The Private Sector Humanitarian Alliance \hspace{1em} $^3$DEVCOM Army Research Laboratory\\
\texttt{\{shal2, jih\}@illinois.edu}}

\begin{document}
\maketitle
\begin{abstract}
Complex news events, such as natural disasters and socio-political conflicts, require swift responses from the government and society. 
Relying on historical events to project the future is insufficient as such events are sparse and do not cover all possible conditions and nuanced situations. Simulation of these complex events can help better prepare and reduce the negative impact. 
We develop a controllable complex news event simulator\footnote{Demo: \url{https://duynguyen2001.github.io/newssimulator/}} guided by both the event schema representing domain knowledge about the scenario and user-provided assumptions representing case-specific conditions.
As event dynamics depend on the fine-grained social and cultural context, we further introduce a geo-diverse commonsense and cultural norm-aware knowledge enhancement component.
To enhance the coherence of the simulation, apart from the global timeline of events,
we take an agent-based approach to simulate the individual character states, plans, and actions. By incorporating the schema and cultural norms, our generated simulations achieve much higher coherence and appropriateness and are received favorably by participants from a humanitarian assistance organization.

\end{abstract}

\input{sections/1intro}
\input{sections/2related_work}

\input{sections/3method}

\input{sections/4exp}

\input{sections/6conclusion}

\section*{Acknowledgement}
This research is based upon work supported by DARPA KAIROS Program No. 18 FA8750-19-2-1004, DARPA SemaFor Program No. HR001120C0123, DARPA CCU Program No. HR001122C0034, DARPA ITM Program No. FA8650-23-C-7316 and DARPA INCAS Program No. HR001121C0165. The views and conclusions contained herein are those of the authors and should not be interpreted as necessarily representing the official policies, either expressed or implied, of DARPA, or the U.S. Government. The U.S. Government is authorized to reproduce and distribute reprints for governmental purposes notwithstanding any copyright annotation therein.

\bibliography{main}
\bibliographystyle{acl_natbib}

\clearpage

\appendix

\input{sections/appendix}

\end{document}

%% file: sections/1intro.tex
\section{Introduction}

%\heng{Technical contributions: let's try to emphasize schema \heng{mention briefly how the schemas are induced from historical events, and cite Zoey's paper} controlled and norm controlled generation}

History repeats itself, sometimes in a bad way, underscoring the importance of recognizing patterns and taking proactive measures to mitigate or ideally eliminate potential natural or man-made disasters. The necessity of this approach is evident in the context of emerging crises such as the COVID-19 pandemic and the Ukraine Crisis. Addressing these situations effectively demands a comprehensive, time-sensitive understanding to inform appropriate decision-making and prompt responses \cite{reddy2024smartbookaiassistedsituationreport}.  
%History repeats itself, sometimes in a bad way. Preventing natural or man-made disasters requires being aware of these patterns and taking pre-emptive action to address and reduce them, or ideally, eliminate them. Emerging events, such as the COVID pandemic and the Ukraine Crisis, require a time-sensitive comprehensive understanding of the situation to allow for appropriate decision-making and effective action response \cite{reddy2024smartbookaiassistedsituationreport}. 
These pressing situations highlight the need for advanced tools capable of scenario simulation to provide predictive insights and facilitate preemptive planning, thereby enhancing preparedness and response strategies.

In developing such a simulator, we define several desiderata: (1) the simulator should be \textbf{controllable}, allowing the user to manage and set the conditions under which the simulation will occur; (2) it must be \textbf{knowledgeable}, meaning it should adhere to and incorporate domain-specific knowledge relevant to the scenario being simulated; (3) the simulator should be \textbf{realistic}, ensuring that each event within the simulation is believable and aligns with commonsense principles; (4) the generated events must be \textbf{coherent}, avoiding any internal conflicts or contradictions; (5) the simulator should exhibit \textbf{sociocultural awareness}, being sensitive to and accurately reflecting diverse geographical contexts and societal norms.

In this context, we introduce \textsc{Miriam}, a novel news event simulator designed to function as an intelligent prophetess.  By leveraging ``What-if'' conditions and assumptions provided by domain experts regarding disaster scenarios, \textsc{Miriam} generates a complex event simulation that describes future events with character-centric narratives, while catering to the geo-cultural diversity inherent in the scenario assumptions. Effectively, our event simulator system that has the following characteristics:
\begin{itemize}[noitemsep]
    \item User-defined assumptions that can steer the direction of the simulation.
    \item Event schemas as input that can be used to constrain the global structure and inject domain knowledge.
    \item Entity-level agent-based simulation which promotes coherence over long simulations.
    \item Norm-aware knowledge enhancement for more culturally appropriate simulations.
\end{itemize}

Figure \ref{fig:overview} shows an overview of \textsc{Miriam}, our proposed system for complex event simulation. By effectively simulating disaster scenarios in both event graph and natural language formats, \textsc{Miriam} aims to assist humanitarian workers and policymakers in conducting reality checks, ultimately aiding in the prevention and management of future disasters.

%% file: sections/2related_work.tex
\section{Related Work}
% \qingyun{I will work on this part}

\begin{figure*}
    \centering
    \includegraphics[width=1\linewidth]{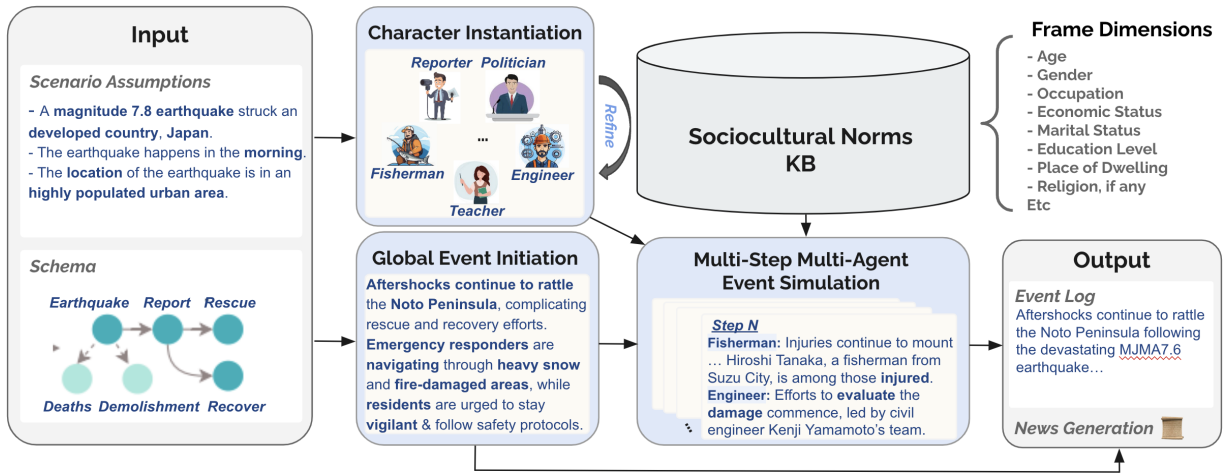}
    \caption{A Simplified Overview of our proposed \textsc{Miriam} System for Complex Event Simulation.}
    \label{fig:overview}
\end{figure*}

\paragraph{Language Model Agents:}
Language models are adept at ``roleplaying'': given the description of a character, the language model can produce responses in character.
Notably, this ability can be used to enable multi-agent collaboration on tasks such as solving logical puzzles~\cite{wang-etal-2024-unleashing}, writing complex software~\cite{Hong2023MetaGPTMP,codeact2024}, reviewing papers~\cite{zeng-2024scientific}, proposing hypothesis~\cite{qi2023large,scimon2024}, machine translation~\cite{bi-etal-2019-multi}, question answering~\cite{puerto-etal-2023-metaqa}, causality explanation generation~\cite{he-etal-2023-lego}, and radiology report summarization~\cite{karn-etal-2022-differentiable}. Another line of work is using LMs to create social simulations~\cite{Suo_2021_CVPR,Park2023GenerativeAI,sun-etal-2023-decoding}, either to improve LM alignment~\cite{Liu2023TrainingSA} or to create synthetic user data for user studies~\cite{Aher2022UsingLL}. However, previous papers concentrate on the feasibility of LM-based social simulation and their alignment with social behaviors. In comparison, we explore using LM agents to assist scenario simulation and story generation. Moreover, unlike existing approaches~\cite{qiu-etal-2022-towards,miceli-barone-etal-2023-dialogue} relying on dialogue to simulate social interactions, our framework generates a comprehensive scenario story that encompasses interactions among various agents, the environment, and the scenario itself. 
\citet{YANG2023117231} conducts a multi-agent simulation to explore residents' consumption behavior under various government regulations.
%by incorporating predefined green consumption utility functions. 
%However, their model limits interactions to binary decision outcomes, either accepting or refusing the policy, whereas our approach allows for a wider range of interactions. 
Our work is also the first to leverage scenario-specific event schemas induced from historical events and culture-specific norms.

\paragraph{Neural Story Generation:}
Due to the complex nature of story generation, controllable story generation has been proposed to address the causality of story events. Existing story generation mainly focuses on two aspects~\cite{goldfarb-tarrant-etal-2020-content}: story planning and character modeling. Previous improvements for story planning can be divided into several categories: keywords planning~\cite{xu-etal-2020-megatron,kong-etal-2021-stylized}, coarse-to-fine planning~\cite{fan-etal-2019-strategies,yao2019-Plan-and-write}, commonsense reasoner~\cite{wang2022contextualized,peng-etal-2022-inferring,peng-etal-2022-guiding}, event graphs~\cite{zhai-etal-2020-story,chen-etal-2021-graphplan,lu-etal-2023-narrative}, and interpersonal relationships~\cite{vijjini-etal-2022-towards}. In contrast, we generate stories in a two-level way, conditioned on event schemas, user-provided assumptions, and commonsense norms. Our work also relates to character modeling in story generation~\cite{Liu2020ACN,zhang-etal-2022-persona}. However, instead of generating character descriptions based on existing stories~\cite{brahman-etal-2021-characters-tell}, we generate character profiles dynamically based on existing events and event schemas. Furthermore, we assign each character as a language agent to simulate his/her interactions with the scenario.

%% file: sections/3method.tex
\section{\textsc{Miriam}: Complex Event Simulator}

%\heng{try to walk through the example in each method subsection}
%\clare{this section is about more than just the components, better to rename it: Approach, and then add a System overview section that walks through one example with Figure 1.}

\subsection{Overview}
Our event simulator takes as input a set of \textbf{assumptions} and an \textbf{event schema}.
Assumptions, provided as free text,  can be scenario-specific, such as the infection rate for disease outbreaks, or scenario-agnostic, such as the (source) location of the event. An event schema is a graph representation of the typical events that occur in a scenario. 
The nodes are atomic events and edges may include temporal edges, hierarchical edges, and logical gates (AND, OR, XOR). The event schema typically encodes prior knowledge about the event scenario (restricting the simulation to parts relevant to the use case). Figure \ref{fig:overview} shows an example of the scenario assumptions and corresponding event schema provided as input to \textsc{Miriam}.

For the output, the system provides the generated simulation in the form of an \textbf{event log} and an \textbf{overview document}.
The event log is a list of event records and profiles of the characters involved in the events. When the event can be grounded to the schema, it has an event type and arguments according to the event ontology.
The overview document is derived from the event log and is a more concise free-text version of the simulation.

\begin{comment}
 \begin{figure*}[t]
     \centering
     \includegraphics[width=\linewidth]{figures/simulator-system.pdf}
     \caption{The overall architecture design of the complex event simulation system. }
     \label{fig:system-design}
 \end{figure*}
\end{comment}
\subsection{System Design: Bi-level Simulation}
\begin{figure}[t]
     \centering
     \includegraphics[width=\linewidth]{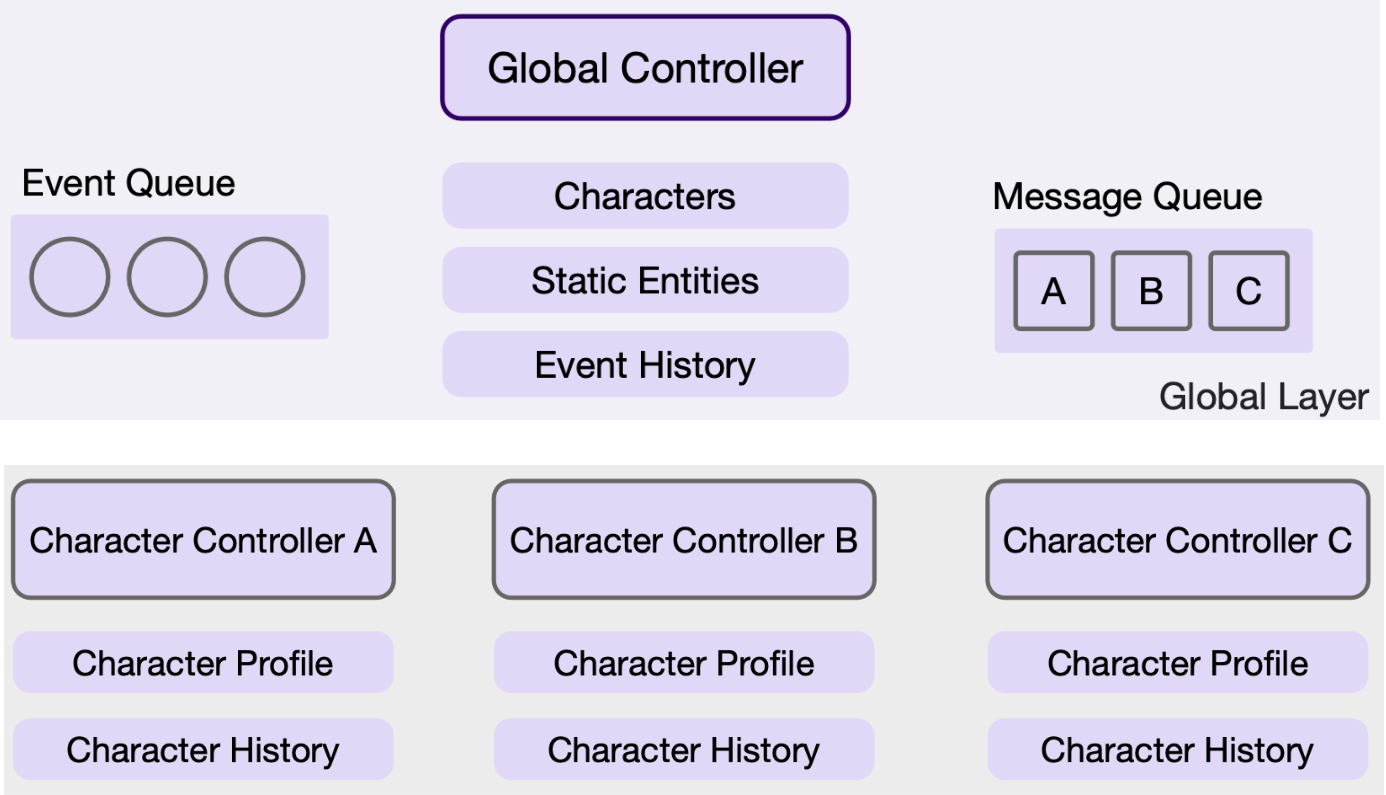}
     \caption{Figure depicting the global and character-level controllers during our simulation generation.}
     \label{fig:bi-level}
 \end{figure}

%We illustrate our system design in Figure \ref{fig:system-design}.
Our simulator contains two levels: the global level and the character level. We will first introduce the two different types of controllers before providing more details (in \S{\ref{sec:sim_events}} and \S{\ref{sec:agent_behavior}}) for the lifecycle of how an event is generated. The global level is defined by the \texttt{Global Controller} object, which takes the event schema and user assumptions as input. We leverage the open-domain schema library induced from our state-of-the-art event schema induction techniques~\cite{hierarchicalschema2023} which covers 41 %\heng{Zoey - can you help me check this number} 
newsworthy scenarios. The global controller maintains pointers to the active characters (their \texttt{Character Controller} objects), entities that have appeared in the simulation, the event history, an event queue, and a message queue. Figure \ref{fig:bi-level} shows the bi-level simulation framework with global and character-level controllers.

The event queue is filled by events from the schema and character controllers. For each time step, once the event queue is filled, the global controller will start to execute the events in temporal order and add the simulated result to the event history.
Message passing in our simulation is implemented by the message queue with the global controller routing the message to the recipient.
The character controllers are more simple in design as they only take care of a single agent. Each controller has its profile and history and is prompted to make plans based on the limited information it acquires.
Initially, there are no character controllers and the characters are generated on-the-fly during the simulation.

\subsection{Simulating Events}
\label{sec:sim_events}
Events go through the cycle of (1) (optionally) event assignment, (2) event planning, (3) event execution, and (4) event reaction.
There are two ways of initiating events, either proposed by the schema or by characters. Events proposed by the schema might undergo the optional event assignment stage, where the \texttt{Schema Event} is assigned to an existing character or creates a new character. This decision is presented to the language model as a multiple-choice question, given the context of the previous simulated events.  Note that some events do not involve any character (such as the mutation of a virus strain) and are directly handled by the global controller. 
In the event planning stage, given the candidate events from the schema, the character controller (or global controller) generates a list of planned events. Each planned event is accompanied by a timestamp that falls between the current time and the next time step of the simulation. This timestamp determines the initial execution order of events but may be affected by event reactions. 
In this step, character controllers also have the liberty of including events not present in the schema. These events will be represented by short text descriptions instead of event types.
Finally, after the planning is complete, each event will be represented as a triple (timestamp, event description, controller name). Figure 3 illustrates the generated planned events from three different controllers. 

% Raw figure is in the last page of this slide deck fyi: https://docs.google.com/presentation/d/1KYDjcfIxMzeFJDcLsg9ptM9bf25HwAcVHppQbOMb4-s/edit?usp=sharing

\begin{figure*}[h!]
    \centering
    \subfloat[Event planning stage.]{
        \includegraphics[width=0.49\textwidth]{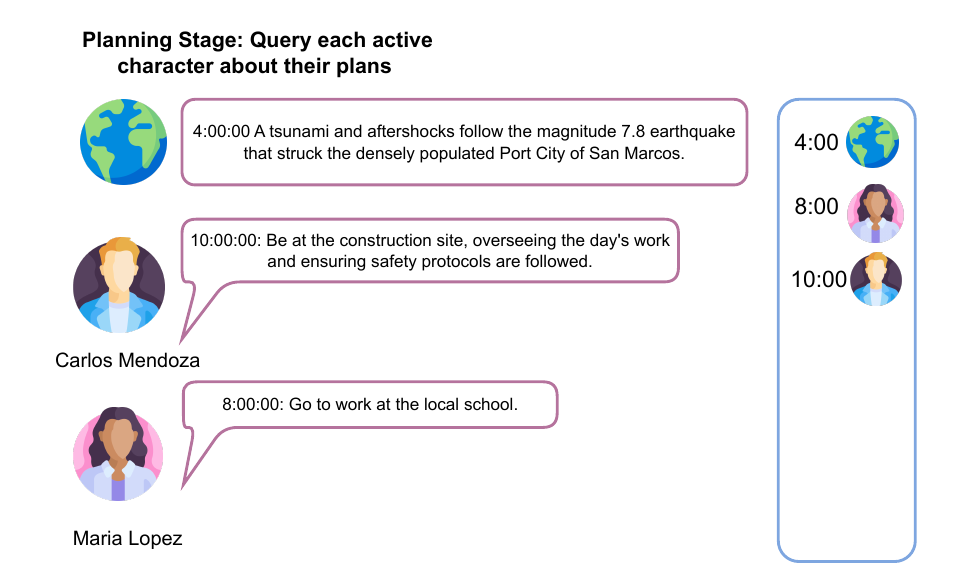}
        \label{fig:planning}
    }
    \subfloat[Event execution and reaction stage]{
        \includegraphics[width=0.49\textwidth]{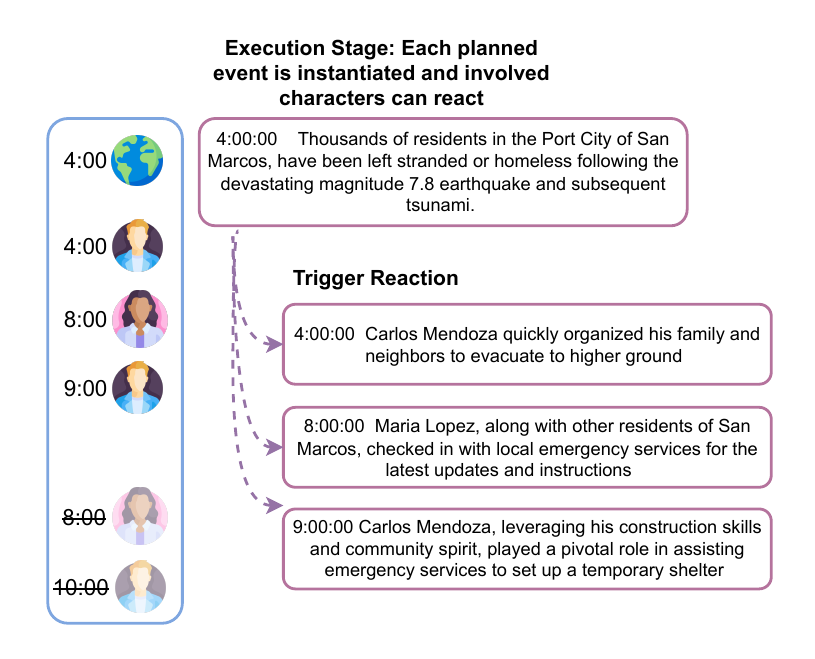}
        \label{fig:execution}
    }
    \caption{In the event lifecycle, the global controller and each one of the characters plans its own events for the next time step. Then all of the plans are centralized and executed in temporal order. If the executed event involves other characters, the other character will be informed and replan its actions.} %\yi{I think it would be too complicated to draw in norms here, so I added in a Fig 1 of the technical workflow and a comparison table. We can position Fig 2 to show the final outcome of our proposed system, for simplicity sake, imo.}
    \label{fig:figure}
\end{figure*}

%\begin{figure*}[h!]
%    \centering
%    \subfloat[Character profile]{
%        \includegraphics[width=0.3\textwidth]{figures/news-simulator-execution.drawio.pdf}
%        \label{fig:execution1}
%    }
%    \subfloat[Event Plan]{
%        \includegraphics[width=0.3\textwidth]{figures/news-simulator-planning.drawio.pdf}
%        \label{fig:planning2}
%    }
%    \subfloat[Event description]{
%        \includegraphics[width=0.3\textwidth]{figures/news-simulator-execution.drawio.pdf}
%        \label{fig:execution3}
%    }
%    \caption{Norm-based revision in the event lifecycle, from character profile to event plan and event description.}
%    \label{fig:figure}
%\end{figure*}

\begin{figure*}[t]
    \centering
    \includegraphics[width=\linewidth]{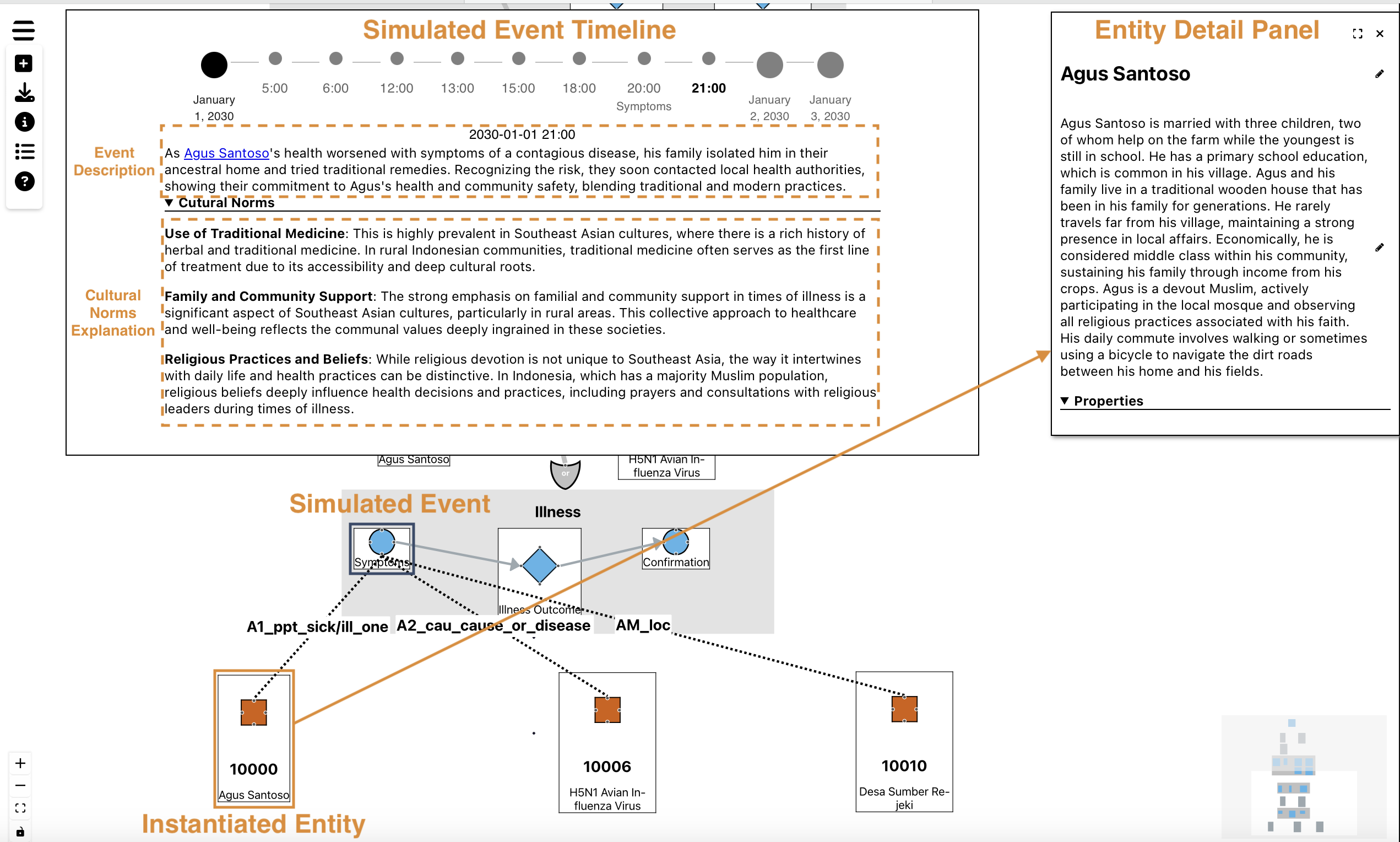}
    \caption{Screengrab of the \textsc{Miriam} interface showing an example simulation for a disease outbreak in Indonesia. The simulation is visualized in the form of an event timeline, with each event provided with a detailed description including related socio-cultural norms, along with background details of the characters involved in the event.}
    %\caption{\heng{need a sharper title, something like Schema and Norm Guided Complex Event Simulation Interface} Simulated H3N2 Decease Outbreak Scenarios in Indonesia visualized using the \textit{Enhanced Schema \& Norm-Driven Complex Event Simulation Interface (eSCNESI).}}
    \label{fig:interface}
\end{figure*}

In the event execution stage, the planned events will be sorted by their timestamp and executed in order. 
Executing an event involves filling in the arguments (including person, location, instrument etc.), and generating a detailed description of the event.
Once executed, the event will be added to the character history and the global event history.

Events that are executed earlier might affect later events. This is handled through reactions (of characters to events).
Concretely, an event proposed by character A but also involves character B will trigger a reaction from character B. Character B can then make alternative plans and change the event queue. For example, A could be a doctor who performs a medical test on a patient B. If the test result is positive, patient B might cancel the remaining plans for the day and become hospitalized. In Figure 3 we see that the two characters Carlos and Maria originally made work plans for their day, but after the execution of the earthquake event, Carlos and Maria react by evacuating and assisting emergency services.

\paragraph{Cultural Enhancement}
%As event simulation is dependent on the geodiverse sociocultural situation, we incorporate sociocultural knowledge across the event simulation pipeline (i.e., character profile, event planning, event description) to enrich the realisticness and insightfulness of the event story generation. Additionally, to cater to a globally interconnected audience, we integrate geodiverse knowledge and sociocultural norm awareness to ensure that the simulated responses are culturally and geographically appropriate, enhancing the global applicability of our simulations. For example, a simulation of an earthquake scenario in Western communities valuing individualism may showcase parents prioritizing their children's safety over their daily professional activities. In contrast, a simulation of an earthquake scenario in China, reflecting communities that generally value collectivism more greatly, may showcase parents first committing to societal rescue efforts before checking on the safety of their own children.
Additionally, event simulation should be dependent on the geodiverse sociocultural situation in order to cater to globally interconnected audience. For example, a simulation of an earthquake scenario in Western communities valuing individualism may showcase parents prioritizing their children's safety over their daily professional activities. In contrast, a simulation of an earthquake scenario in China, reflecting communities that generally value collectivism more greatly, may showcase parents first committing to societal rescue efforts before checking on the safety of their own children. To address this, Miriam integrates sociocultural knowledge across the event simulation pipeline to enrich the realisticness and insightfulness of the event story generation, as well as to ensure that the simulated responses are culturally appropriate.  %(i.e., character profile, event planning, event description) 
%\heng{I suggest to move the following paragraph to technical method} Additionally, to cater to a globally interconnected audience, Miriam integrates geodiverse knowledge and sociocultural norm awareness to ensure that the simulated responses are culturally and geographically appropriate, enhancing the global applicability of our simulations. For example, a simulation of earthquake scenario in western communities valuing individualism may showcase parents prioritizing their children safety over their daily life professional activities. In contrast, a simulation of earthquake scenario in China, reflecting communities that generally more greatly value collectivism, may showcase parents first committing to societal rescue efforts before checking on the safety of their own children. %\heng{add a new figure to show the norm based revision}

\begin{itemize}[leftmargin=*,topsep=0.8pt]
    \item \textbf{Character Profile Initialization:} When simulating event scenarios, the fine-grained background information (e.g., age, gender, occupation, marriage/family status, economic status, education level, ethnicity, religious beliefs, etc.) of each simulated individual really matters, but an LLM may often miss important social profile dimensions while generating the initial character descriptions. We leverage the social theory grounded formulation in~\citet{ziems-etal-2023-normbank} and ask LLM to enhance the initial character profile descriptions for any important missing social profile dimensions.
    \item \textbf{Per-Character Event Description:} To better tailor event descriptions towards the cultural norms of a particular society being simulated, we leverage the concept of norm discovery on-the-fly ~\cite{fung-etal-2023-normsage}. Specifically, we discover relevant social norms through LLM self-retrieval augmented generation grounded on the concept of internal knowledge elicitation, and further supplement the norms with the set of pre-existing norms from~\citet{fung2024massively}, which covers massively multi-cultural norm for 1000+ sub-country regions and 2000+ ethnolinguistic groups (discovered through web documents via ShareGPT), to dynamically construct and enrich the NormKB relevant for the scenario context. Then, we rank the social norms by relevance and insightful to the situation context, and condition on these social norms as additional context for refining news simulation with greater cultural detail. Specifically, we refine the event descriptive by a LLM prompting mechanism that takes as input the original event description, as well as the relevant sociocultural norms for auxilliary context, followed by the task instruction of \textit{"Revise the event description to be more tailored to the unique cultural norms, while keeping the overall event description a similar length"}, to derive the norm-enhanced event description. %For example, \heng{Yi todo: add how refinement is done and give examples; add examples from slides, maybe organize them into a table}. 
    %"Event Description, Elaborated:\n" + evt_desc + "\n\n" + \
    %"Sociocultural Norms:\n" + evt_based_norms + \
    % "\n\n---\n\n" + "Revise the event description to be more tailored to the unique cultural norms, while keeping the overall event description a similar length:" evt_desc, _ = call_OpenAI_api(prompt_in, max_gen_len=2048)
\end{itemize}

We refer the reader to Table \ref{tab:tab1} in the appendix for an example comparing event simulation with and without cultural norm enhancement.

\subsection{Simulating Agent Behavior}
\label{sec:agent_behavior}
We can also inspect our simulation on a character level. 
Each character in the system is defined by a name, age, profession, backstory, and plotline.
Different from prior work, the characters in our system are created dynamically by the global controller. 
The attributes of the character are generated upon creation based on the global assumptions and the event that the character participates in. 

% \clare{It would be helpful, in pivoting here to discuss agents, to spell out the relation/contrast between characters and agents. I'm not clear about the distinction between active characters and active agents. Consider also adding in a footnote to explain that ``actor'' here refers to a linguistic argument of an event}
At the beginning of each time step of the simulation, every active character(agent) will be polled for their upcoming planned events. 
Each character will keep track of the events that he/she has been involved in. These memories will be part of the input when the agent makes up the plan.

In particular, we introduce a \textbf{self-critique loop} to the planning stage. The theory-of-mind inspires this self-critique loop: the model is required to infer what the agent will do so that the plot is fulfilled (while the agent does not know about the plot). To model this second-order relationship, we first ask the model to role-play as the character and generate a draft plan based on the character profile and character history. Then the model is instructed to behave as a critic and check if the actions agree with the plot. 
The critic will give detailed feedback on which actions should be kept, removed, or revised, along with the reasoning for adjustments (``you are feeling unwell today so you should not go out'').
In our system, this self-critique will stop when the critic does not have any suggestions or when we reach a maximum of 3 rounds.

Since the efficiency of the simulation is heavily influenced by the number of active agents, we set a threshold for the maximum number of characters active at each time step. If the current number of characters exceeds that threshold, we retire the least recently used character.

%% file: sections/4exp.tex
\section{Experiments}
\label{sec:expts}
Our experiments aim to investigate the impact of various components integrated into our system, alongside assessing the overall utility of the tool. First, \S{\ref{sec:evaluation_setup}} outlines the automatic evaluations to determine the benefit of leveraging the event schemas and cultural norms in simulation generation. Then, \S{\ref{sec:utility_eval}} studies the perceived utility of our tool, based on feedback from participants affiliated with a humanitarian assistance organization. The \textsc{GPT-4o mini} model serves as the underlying LLM in the simulation generation process.
%We perform both human and automatic evaluations on the generated simulations.
%\clare{what is/are the research question/s here that the experiments are to address? is it, what is value added of the schemas, so comparing LLM-baseline approach vs LLM-with-schema approach? And/Or is it, what is the value added of LLMs, so comparing traditional schema-only methods vs LLM-agent-based approach?}

\subsection{Automatic Evaluation}
\label{sec:evaluation_setup}

To demonstrate the benefit of incorporating the event schemas and cultural norms into our system, Table \ref{tab:auto_eval} presents a comparative analysis of simulations generated by different variants of our system. Our approach, designated as \textit{Schema + Norms}, is evaluated against (a) \textit{Schema Only}, which does not utilize cultural norms, and (b) \textit{W/O Schema}, which employs the LLM directly to generate simulations without schema guidance. The evaluation criteria include a range of metrics: (i) \textit{coherence}, assessing the overall flow of the simulation, (ii) \textit{entailment}, determining whether the simulation aligns with given assumptions, (iii) \textit{realism}, evaluating the plausibility of the simulation in the given scenario, and (iv) \textit{cultural appropriateness}. We employed \textsc{GPT-4o} for the automatic evaluation of simulation quality, with detailed prompts provided in Table~\ref{table:eval_prompts} in the Appendix. The evaluation covered 47 simulations generated for scenarios including 'Earthquake,' 'Disease Outbreak' and 'Chemical Spill,' across five distinct regions (cultures): the United States, France, China, Peru, and Indonesia. The results demonstrate that incorporating both cultural norms and event schemas significantly enhances the quality of the generated simulations across all metrics, with notable improvements in cultural appropriateness and entailment with assumptions.

\begin{table}[!htb]
    \centering
    \small
    \def\arraystretch{1.5}
    \begin{tabular}{c|c|c|c}
    \toprule
    \multicolumn{1}{c|}{\multirow{2}{*}{\textbf{Metric}}} & \textbf{Schema +} & \textbf{Schema} & \textbf{W/O}\\
    & \textbf{Norms} & \textbf{Only} & \textbf{Schema} \\
    \midrule
    Coherent & \textbf{7.49} & 6.94 & 6.57\\
    Entailment & \textbf{8.36} & 8.11 & 7.23\\
    Realistic & \textbf{7.61} & 7.09 & 6.79\\
    Appropriate & \textbf{8.57} & 7.02 & 6.60 \\
    \bottomrule
    \end{tabular}
    \caption{Automatic evaluation (rated by GPT-4o) of simulations generated by different variants of our system.}
    \label{tab:auto_eval}
\end{table}
\begin{comment}
\begin{table}[!htb]
    \centering
    \small
    \def\arraystretch{1.5}
    \begin{tabular}{c|c|c|c|c}
    & \multicolumn{1}{c|}{\multirow{2}{*}{\textbf{Metric}}} & \textbf{Schema +} & \textbf{Schema} & \textbf{W/O}\\
    & & \textbf{Norms} & \textbf{Only} & \textbf{Schema} \\
    \hline
    \parbox[t]{2mm}{\multirow{3}{*}{\rotatebox[origin=c]{90}{\textbf{Simulation}}}} & Coherence & \textbf{7.9} & 5.8 \\
    & Entailment & \textbf{9.1} & 6.8\\
    & Relevance & \textbf{9.0} & 6.6\\
    \hline
    \parbox[t]{2mm}{\multirow{3}{*}{\rotatebox[origin=c]{90}{\textbf{Character}}}} & Coherence & & \\
    & Entailment & & \\
    & Relevance & & \\
    \hline
    \end{tabular}
    \caption{Comparison of simulations generated with schema vs without when rated by GPT-4o (on a scale of 1-10) at simulation-level and character-level for different metrics.}
    \label{tab:auto_eval}
\end{table}
\end{comment}

\subsection{Human Utility Evaluation}
\label{sec:utility_eval}
\begin{figure}[t]
    \centering
    \includegraphics[width=\linewidth]{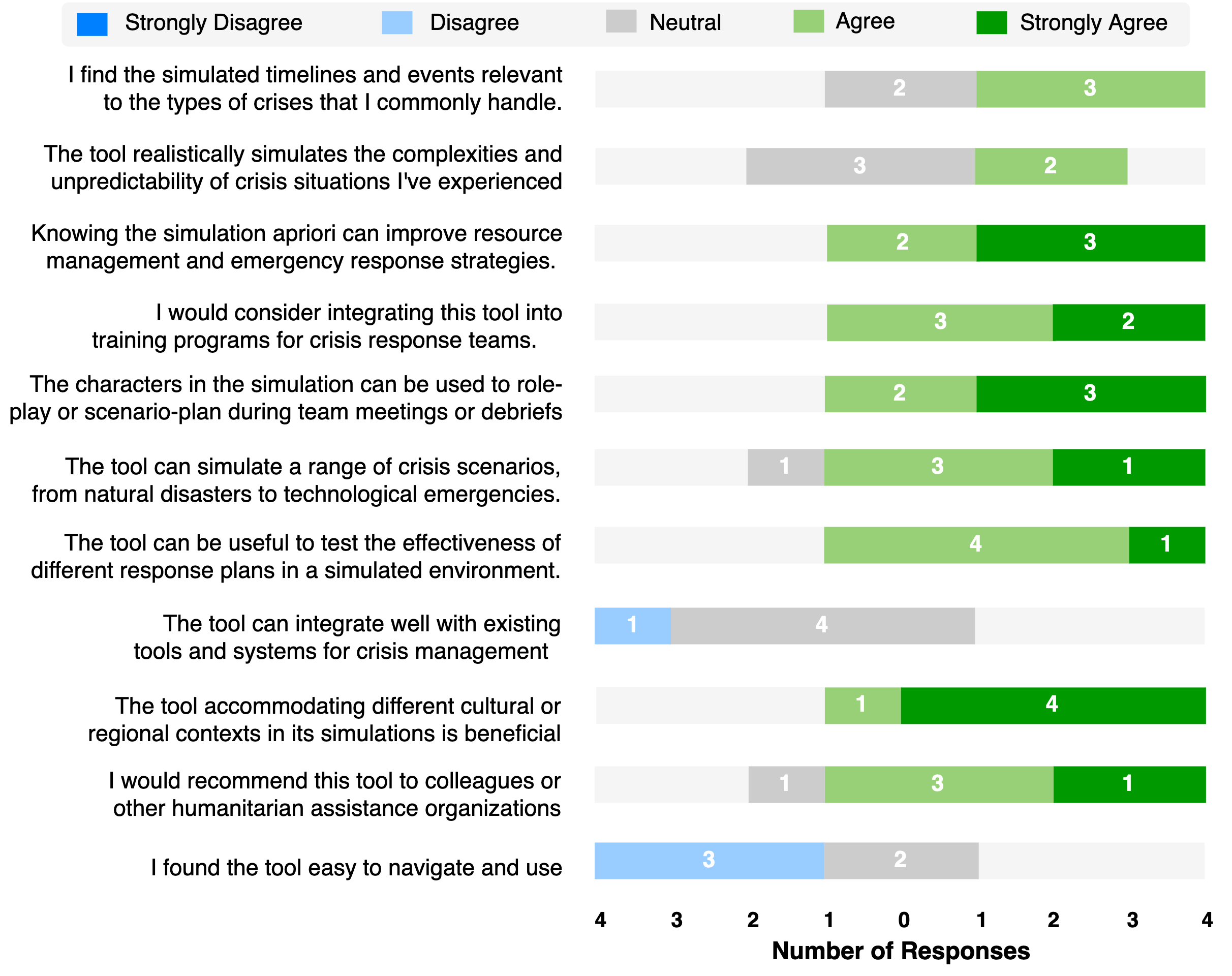}
    \vspace{-1.5em}
    \caption{Results from utility evaluation by participants from a humanitarian assistance organization.}
    \label{fig:utility_eval}
    \vspace{-1em}
\end{figure}
We conducted a human evaluation to assess the perceived utility of the tool. The study involved five participants from a humanitarian assistance organization who navigated the generated simulations using the interface depicted in Figure \ref{fig:interface}. Participants provided qualitative feedback during the study and completed a post-study questionnaire for quantitative evaluation. The results, presented in Figure \ref{fig:utility_eval}, indicate that participants found the system promising and useful for training crisis response teams. However, feedback highlighted significant areas for improvement in the interface, suggesting that the current version may be limiting when integrating the system into existing workflows. We plan to address these issues in future iterations based on the qualitative feedback received.

%% file: sections/6conclusion.tex
\section{Conclusion}
We introduce \textsc{Miriam}, a controllable complex news event simulator designed to improve preparation and response to events like natural disasters and socio-political conflicts. Using event schemas for domain knowledge and incorporating user assumptions, Miriam offers global control over event dynamics. It enhances realism by integrating geo-diverse commonsense and cultural norm awareness. The system generates a coherent global timeline and employs a large language model to simulate the states, plans, and actions of individual agents, enabling detailed and realistic character-based stories. This agent-based approach outperforms traditional schema-only methods, providing a valuable tool for training, preparedness, and societal resilience.

\begin{comment}
\clare{Other comments: \\
(1) LLMs:  the readers will want to know, somewhere in section 3 about the system, which LLM is being used in which components. (I do see the mention of GPT-4o in automated evaluation section, but that is for evaluating the system output.)\\
(2) Table 1: since all 3 categories (profile, event plan, event description) are per-character, is there a way to show these 3 categories for the same character?\\
(3) Figure 1: I was expecting that the scenario assumptions (top of input box) would also feed with the Schema into the Global Event Initialization box. Would it make sense to add an arrow for that?\\
(4) Input \& user: what is the source of assumptions? Does the domain expert spell these out, or are they pre-populated in system for proof of concept? Does the domain expert select the schema from the schema library, is the schema also provided with the pre-populated assumptions?\\
(5) Fig 1 \& Table 2: what would be the workflow in Fig 1 corresponding to the "without" schema runs in Table 2?\\
(6) What parts of Fig 1 correspond to the simulation \& character content evaluated in Table 2?}
\end{comment}

%% file: sections/appendix.tex
\begin{table*}[ht]
    \begin{tabular}{p{1.5cm}|p{4cm}|p{9.2cm}}
        \hline
         &  \textbf{w/o Norms} & \textbf{w/ Norms} \\
         \hline 
        \textbf{\small{character \newline profile}} & \small{Li Wei is a 32-year-old marketing executive living in Shanghai. He is known for his outgoing personality and enjoys attending social events and networking gatherings. Li Wei is health-conscious but often finds himself in crowded places due to his job.} & \small{Li Wei is a 32-year-old marketing executive living in Shanghai.  He is known for his outgoing personality and enjoys attending social events and networking gatherings. Li Wei is health-conscious but often finds himself in crowded places due to his job. \textit{\textbf{He is single and lives in a modern high-rise apartment in the bustling Jing'an District of Shanghai. He holds a MBA from Fudan University. Originally from a smaller city in Jiangsu province, he moved to Shanghai ten years ago to pursue his career. Economically, Li Wei is well-off, earning a comfortable salary that allows him to indulge in his interests and maintain a cosmopolitan lifestyle. He usually commutes to work primarily by metro, which is efficient and fits his environmentally conscious values. Although culturally rooted in Confucian values, Li Wei is not particularly religious, focusing more on personal and professional growth.}}} \\
        \textbf{\small{per-character event plan}} & \small{Hispanic Single-Mother Amidst Covid Outbreak in the US: wake up early to get prepared for morning shift as part-time shopping mall cashier. Check news and see gov't announces pandemic lockdown, which causes her shift to be cancelled. Prepares healthy breakfast for daughter and helps her prepare for remote class.} & \small{Hispanic Single-Mother Amidst Covid Outbreak in the US: wake up early to get prepared for morning shift as \textbf{part-time} shopping mall cashier. Check news and see gov't announces pandemic lockdown, doesn't have a job now and searches for \textbf{gov't subsidy} options. Prepares healthy omelette breakfast for daughter and helps her prepare for remote class over \textbf{zoom}.} \\
        \textbf{\small{per-character event \makecell{description}}} & \small{Unwind at Home: Despite the ongoing outbreak in Jakarta, Andi Pratama decided to go for a morning jog in the park, taking extra precautions to avoid crowded areas and maintain personal hygiene.} & \small{Unwind at Home: During a disease outbreak in Jakarta, Andi Pratama, a devout Muslim, performed the \textbf{Tahajjud prayer} at night in his apartment. As the new year began, he prayed earnestly for his community's well-being. In the morning, after performing \textbf{Fajr prayer} at home, Andi Pratama jogged in a nearby park, embracing the \textbf{"gotong royong" spirit} by carefully avoiding crowded areas and keeping distance from others.} \\
        \hline
    \end{tabular}
    \caption{Comparison of event simulations with and w/o knowledge enhancement from culture-specific social norms.}
    \label{tab:tab1}
\end{table*}

\begin{table*}[ht]
\begin{center}
%\begin{adjustbox}{width=1.0\textwidth}
\begin{tabular}{m{15cm}}
\toprule

\tiny{
    \texttt{\textbf{\hspace{30em}\underline{Evaluation Prompt}} \newline \newline
You are an automatic quality evaluator. You will be provided with some simulations and you will need to evaluate them based on the criteria that is mentioned.\newline
---\newline
You are provided with some simulations corresponding to the scenario: \{scenario\_name\}
\newline
---\newline
The simulations were generated based on the following assumptions in no specific order:
\newline
---\newline
Assumptions: \{list\_of\_assumptions\}
\newline
---\newline
The simulations are below. Each simulation is from the future in the form of a listwise log of events. Each log item has the time and a description of the event.
\newline
---\newline
Simulation 1: \{list\_of\_events\}
\newline
---\newline
Simulation 2: \{list\_of\_events\}
\newline
---\newline
Simulation 3: \{list\_of\_events\}
\newline
---\newline
Metric: For each of the simulations, you need to evaluate how coherent the simulation is and provide a single score in the range of 1-10, where a higher score indicates better coherence.
\newline
---\newline
DO NOT bias your judgment based on the length of the simulation. You should only respond in the JSON format as described below. You SHOULD ensure that the provided output can be directly parsed into json using python json.loads
\newline
---\newline
Response Format:
\newline
\{\{
\newline
"thoughts": "Your step-by-step reasoning for the evaluation scores you will provide",\newline
"simulation\_1": "Score for simulation 1. Just provide a number here in the range of 1 to 10",\newline
"simulation\_2": "Score for simulation 2. Just provide a number here in the range of 1 to 10",\newline
"simulation\_3": "Score for simulation 3. Just provide a number here in the range of 1 to 10",\newline
\}\}
}}
\\
\bottomrule
\end{tabular}
%\end{adjustbox}
\caption{Prompts for automatic evaluation of the simulations.}
\label{table:eval_prompts}
\end{center}
\end{table*}